\documentclass[letterpaper, 10 pt, conference]{ieeeconf}  

\usepackage{cite}
\usepackage{amsmath,amssymb,amsfonts}
\usepackage{algorithmic}
\usepackage{graphicx}
\usepackage{textcomp}
\usepackage[dvipsnames]{xcolor}
\usepackage{multirow}
\usepackage{makecell}
\usepackage{tikz}
\usepackage{subcaption}
\usepackage{color, colortbl}
\usepackage{xcolor}
\usepackage{tikz}
\usepackage{url}
\usepackage{soul}
\usepackage{hyperref}
\usepackage{lipsum}

\usetikzlibrary{arrows.meta}

\definecolor{LightBlue}{RGB}{212, 250, 252} 

\def\mycircle[#1]{\tikz\draw[#1,fill=#1] (0,0) circle (.5ex);}

\def\arrowright[#1]{\begin{tikzpicture}
  \draw[#1, -{Triangle[width = 5pt, length = 2pt]}, line width = 2pt] (0.0, 0.0) -- (0.2, 0.0);
\end{tikzpicture}}

\def\arrowleft[#1]{\begin{tikzpicture}
  \draw[#1, -{Triangle[width = 5pt, length = 2pt]}, line width = 2pt] (0.2, 0.0) -- (0.0, 0.0);
\end{tikzpicture}}

\def\arrowup[#1]{\begin{tikzpicture}
  \draw[#1, -{Triangle[width = 5pt, length = 2pt]}, line width = 2pt, rotate=270] (0.2, 0.0) -- (0.0, 0.0);
\end{tikzpicture}}

\def\arrowstraightleft[#1]{\begin{tikzpicture}
  \draw[#1, -{Triangle[width = 5pt, length = 2pt]}, line width = 2pt] (0.2, 0.0) -- (0.0, 0.0); 
  \draw[fill=#1,#1] (0.15,0) rectangle ++(0.05,-0.15);
\end{tikzpicture}}

\IEEEoverridecommandlockouts                              

\newcommand\copyrighttext{%
  \footnotesize \textcopyright 2024 IEEE. Personal use of this material is permitted.
  Permission from IEEE must be obtained for all other uses, in any current or future
  media, including reprinting/republishing this material for advertising or promotional
  purposes, creating new collective works, for resale or redistribution to servers or
  lists, or reuse of any copyrighted component of this work in other works.}
\newcommand\copyrightnotice{%
\begin{tikzpicture}[remember picture,overlay]
\node[anchor=south,yshift=10pt] at (current page.south) {\fbox{\parbox{\dimexpr\textwidth-\fboxsep-\fboxrule\relax}{\copyrighttext}}};
\end{tikzpicture}%
}

\title{\LARGE \bf
TLD-READY: Traffic Light Detection - Relevance Estimation and Deployment Analysis
}

\author{Nikolai Polley$^{1}$, Svetlana Pavlitska$^{1,2}$,  Yacin Boualili$^{1}$, Patrick Rohrbeck$^{1}$, Paul Stiller$^{1}$,  \\ Ashok Kumar Bangaru$^{1}$, and J. Marius Zöllner$^{1,2}$
\thanks{$^{1}$ Karlsruhe Institute of Technology (KIT), Kaisterstr. 12 Karlsruhe Germany. {\tt\small \{prename.surname\}@kit.edu}.}
\thanks{$^{2}$ Department of Technical Cognitive Systems, FZI Research Center for Information Technology, Germany.
	{\tt\small \{surname\}@fzi.de}.}%
}

\begin{document}

\maketitle
\copyrightnotice
\thispagestyle{empty}
\pagestyle{empty}

\begin{abstract}
Effective traffic light detection is a critical component of the perception stack in autonomous vehicles. This work introduces a novel deep-learning detection system while addressing the challenges of previous work.
Utilizing a comprehensive dataset amalgamation, including the Bosch Small Traffic Lights Dataset, LISA, the DriveU Traffic Light Dataset, and a proprietary dataset from Karlsruhe, we ensure a robust evaluation across varied scenarios. 
Furthermore, we propose a relevance estimation system that innovatively uses directional arrow markings on the road, eliminating the need for prior map creation. On the DriveU dataset, this approach results in 96\% accuracy in relevance estimation. Finally, a real-world evaluation is performed to evaluate the deployment and generalizing abilities of these models. 
For reproducibility and to facilitate further research, we provide the model weights and code: \url{https://github.com/KASTEL-MobilityLab/traffic-light-detection}.

\end{abstract}

\section{INTRODUCTION}
Accurate and reliable detection of traffic light signals is essential for autonomous vehicles at Level 3 autonomy and above. Moreover, even Advanced Driver-Assistance Systems (ADAS) at Level 2 autonomy can derive substantial benefits from enhanced detection capabilities, potentially alerting inattentive drivers of phase changes. While some strategies leverage Vehicle-to-Everything (V2X) communication to acquire traffic signal phase information from intelligent infrastructure~\cite{ochs2024one, zipfl2020traffic}, the predominant body of research focuses on real-time camera-based detection methodologies. Here, learning-based neural network techniques have become increasingly prevalent, largely superseding traditional image-processing techniques, especially for the localization of traffic lights. Despite the availability of numerous datasets, we show that none offer a comprehensive set of all required information. 
Furthermore, much of the research relies on in-house, private datasets, and a lack of standardized evaluation metrics leads to a fragmented research landscape in which open-source implementations are considerably scarce~\cite{pavlitska2023traffic}.
An often neglected part of traffic light detection 
is the assignment of signal relevance to the ego vehicle. It is common for multiple traffic light signals, corresponding to different lanes, to be visible simultaneously; thus, it is crucial for the system to distinguish between signals that are relevant to the ego vehicle and those that are not. On these remarks, our contributions can be summarized as follows:
\begin{itemize}
    \item We evaluate existing traffic light detection datasets and apply state-of-the-art (SOTA) object detection models to three public and one proprietary dataset, making model weights and code available open-source.
    \item We introduce a novel methodology for assigning relevance scores to traffic signals by leveraging lane markings as auxiliary indicators.        
    \item We evaluate the proposed models and approaches with an automated vehicle in real traffic in the Test Area Baden-Württemberg in Germany. We analyze the challenges that occur when deploying models trained on publicly available datasets on public roads. We describe the limitations of existing datasets and propose mitigation methods.
\end{itemize}

\begin{figure}[t]
\centering
\begin{subfigure}[t]{\linewidth}
    \includegraphics[width=\textwidth]{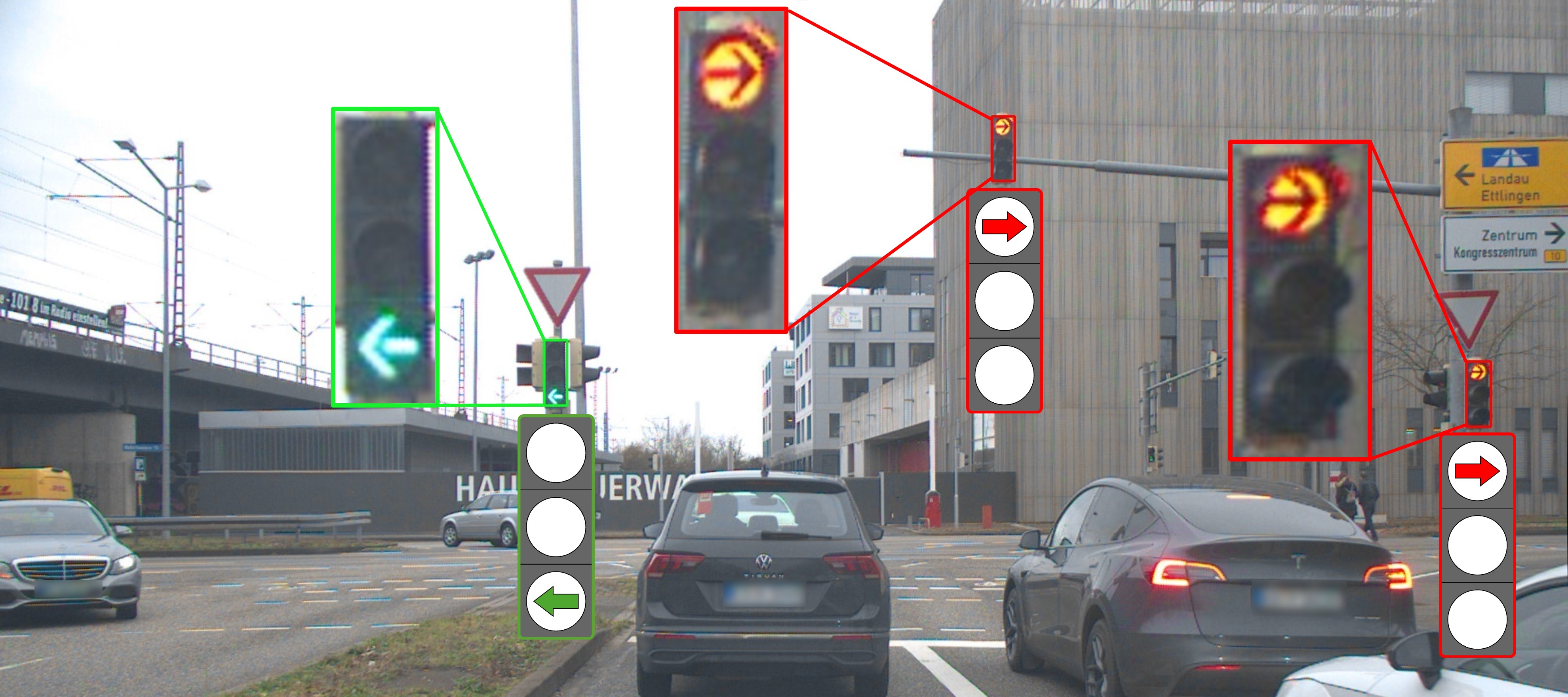}
\end{subfigure}
\begin{subfigure}[t]{\linewidth}
    \includegraphics[width=\textwidth]{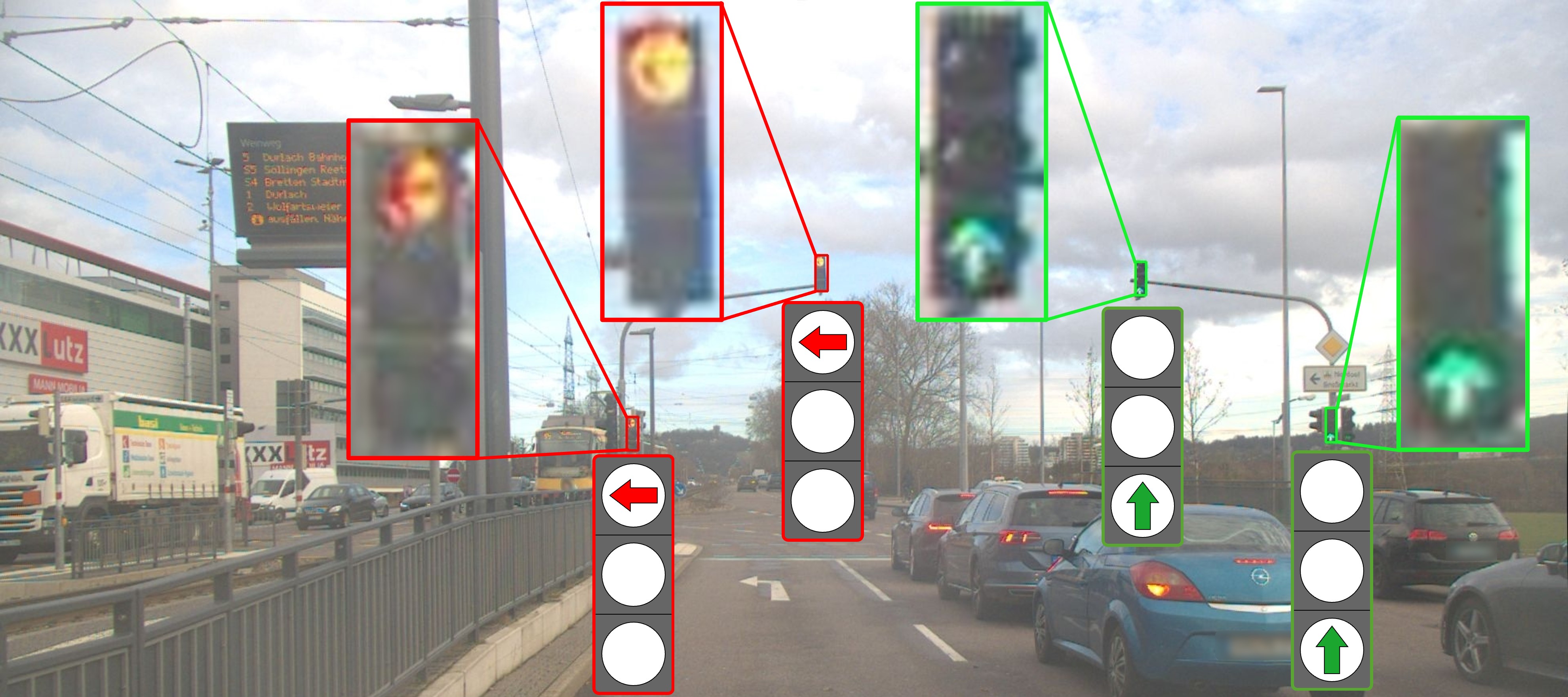}
\end{subfigure}
\begin{subfigure}[t]{\linewidth}
    \includegraphics[width=\textwidth]{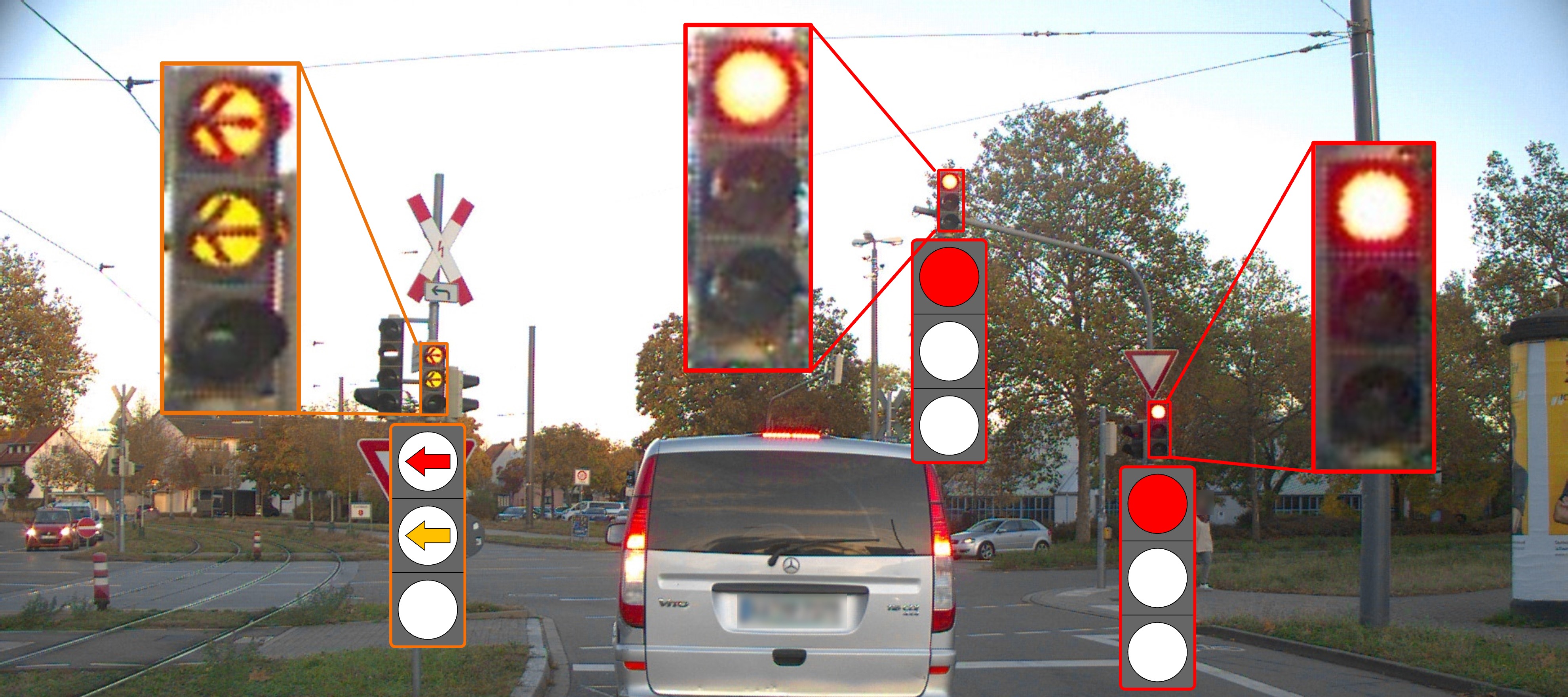}
\end{subfigure}

    \caption{DTLD-YOLOv8x model during a test drive in Karlsruhe. For clarity, close-ups of traffic lights are added, and icons are used to visualize the predictions of the model. }
    \label{fig:concept}
\end{figure}

\clearpage
\newpage

\section{RELATED WORK}
\subsection{Traffic Light Detection}
As this work focuses exclusively on camera-based methods for traffic light detection, Vehicle-to-Everything (V2X) communication methods are excluded from consideration.
Early approaches predominantly relied on manually defined feature detectors, sometimes complemented by machine-learned classifiers~\cite{chung2002vision, lindner2004robust, nienhuser2010visual}. These methods lacked the capability to operate in real-time~\cite{lindner2004robust, chung2002vision} and did not provide robust traffic light detection under varied conditions, e.g., Nienhuser~\cite{nienhuser2010visual} reports true positives if only a single frame in a sequence of frames is correctly classified. 
More recently, researchers have predominantly adopted deep learning-based approaches, primarily employing convolutional neural networks~\cite{pavlitska2023traffic}. Typically, the task involves localizing and classifying a variable number of traffic light instances within a single image. While the result of the localization task is mostly approached similarly, namely as 2D-bounding boxes for each instance, the classification task differs greatly for each approach. While some works only differentiate between \texttt{stop} and \texttt{go} 
~\cite{john2014traffic, john2015saliency, possatti2019traffic, sanitz2023small}, other works classify into three or more classes representing the states of the traffic lights, \texttt{red, yellow, green}. A smaller number of works also classify the traffic light's pictograms, mostly arrows referencing corresponding lanes. The underlying architectures of these works can be clustered into three categories (cf.~\cite{pavlitska2023traffic}): The most used category,  \textit{modification of generic object detectors}, employs object detectors like YOLO~\cite{redmon2016you, redmon2018yolo9000, redmon2018yolov3, bochkovskiy2020yolov4}, SSD~\cite{liu2016ssd}, or Faster-RCNN~\cite{ren2015faster}, which are retrained specifically for traffic light detection~\cite{aneesh2019real, bach2018deep, gokul2020comparative, jensen2016vision, liu2023traffic, muller2018detecting, pon2018hierarchical, yan2021end}. Occasionally, these generic architectures are adapted for the task at hand, usually incorporating enhancements for detecting small objects~\cite{han2019real, wang2022traffic, naimi2021fast}. Architectural designs, which go beyond conventional object detectors, can be categorized into \textit{single-stage} and \textit{multi-stage} approaches. \textit{Single-stage} approaches simultaneously localize and classify traffic lights in a unified process \cite{weber2016deeptlr, weber2018hdtlr, john2014traffic, john2015saliency}. Conversely, \textit{multi-stage} approaches separate the localization and classification tasks into distinct phases. Typically, a generic object detector is utilized as a region of interest (ROI) extractor, followed by a specialized classification CNN~\cite{lu2018traffic, wang2018method, kim2019traffic, jayasinghe2022towards, choi2024real}. Alternatively, some methods employ conventional detectors paired with non-deep learning techniques for classification~\cite{kim2018deep, yudin2018usage, gupta2019framework, tran2020accurate, nguyen2020robust}. Zhang et al.~\cite{zhang2023robust} describe a multi-stage framework using a pre-trained YOLOv8~\cite{jocher2023yolov8} model for initial detection, followed by a rule-based HSV thresholding filter for state classification. Additionally, a fine-tuned YOLOv8 is employed for classification between four types of pictograms. They evaluate their approach on the Tongji Small Traffic Light Dataset (TSTLD) dataset, containing approx. 1,500 images featuring approx. 3,000 traffic light instances and achieve in this dataset high accuracies of 95\%.
Recently, some works have replaced convolutional neural networks with transformer-inspired~\cite{vaswani2017attention} and DETR-like~\cite{carion2020end} models and apply these new methodologies to traffic light detection~\cite{greer2023robust, ou2022traffic}. However, these architectures are unable to operate in real-time due to inherent performance constraints of their underlying architecture.  Chuang et al.~\cite{chuang2023traffic} propose an object detector consisting of a feature extraction backbone of an E-ELAN-YOLOv7~\cite{wang2023yolov7} architecture with residual bi-fusion (PRB) feature pyramid networks and attention modules. The system was evaluated using the Bosch Small Traffic Light Dataset (BSTLD), achieving a mean Average Precision (mAP) at  0.5 IoU of 0.665; however, the study does not address the model's real-time performance capabilities.
Until now, no transformer architecture capable of real-time performance, such as RT-DETR\cite{lv2023detr}, has been applied to traffic light detection.

A recent survey stresses the difficulty in comparing these different methodologies, as datasets, tasks, hardware resources, and evaluation metrics differ~\cite{pavlitska2023traffic}. Only a select number of models provide open-source code and none of these are able to classify pictograms.

\subsection{Relevance Estimation}
An autonomous vehicle must not only detect traffic lights but also distinguish between those it must obey and those that are irrelevant as they are responsible for other driving lanes.

The open-source autonomous driving platform Apollo~\cite{apollo} stores world coordinates of traffic lights in HD-Maps and projects them into images using the localization module of the vehicle. As the vehicles' calibration, localization, and HD-Maps aren't exact, the projection is unreliable and requires large ROI zones. The main disadvantage of Apollo and similar map-based-projection approaches~\cite{fairfield2011traffic,levinson2011traffic, john2015saliency, possatti2019traffic} is the requirement of prior-maps which restricts their functionality to annotated streets only. Particularly scarce are annotations that encompass not only the position of traffic light instances but also their mounting heights, both of which are essential for this projection approach. 

A different approach trains a model with a modified loss term, rewarding the detection of relevant traffic lights~\cite{greer2023robust}. Although their recall increases for relevant traffic lights, their approach cannot distinguish between relevant and irrelevant traffic lights. 

Langenberg et al.~\cite{langenberg2019deep} adopt a deep fusion approach utilizing the positions of traffic lights, lane line markings, and lane arrows as inputs to a convolutional neural network, which returns a column-like location of all relevant traffic lights. However, these inputs are human-annotated ground truths that are generally not accessible in real-world driving. 

In another approach, a rule-based algorithm detects all traffic lights and marks the largest and topmost traffic light as relevant~\cite{li2017traffic}. No evaluation is performed, and we assume inadequate quality as the top-most largest traffic lights are, in a multitude of cases, not relevant for the ego vehicle. 

Trinci et al.~\cite{trinci2023cross} employ a combination of two CNN architectures to estimate the relevance of traffic lights in consecutive images. They frame the problem as a pure classification task, determining whether the vehicle should stop, and evaluate this using the $AP_{stop}$ metric. Detection of multiple traffic lights and relevance estimation of other states is not addressed.

\clearpage
\newpage

\section{DATASETS AND LABELS}

Out of available traffic light detection datasets, we omit those with a relatively small number of images, including \texttt{LaRa}~\cite{LARA} (11K images, four classes), \texttt{WPI}~\cite{chen2016accurate} (3K images, 21 classes),  \texttt{Cityscapes TL++}~\cite{janosovits2022cityscapes} (5K images, 6 classes), \texttt{S$^2$TLD}~\cite{yang2022scrdet++} (5K images, 5 classes), and \texttt{DualCam}~\cite{jayarathne2023dualcam} (1k images, 10 classes). We evaluate models on four datasets (see Table~\ref{tab:datasets}). Most datasets only provide \texttt{train/val} split. To ensure evaluation on unseen data, we split the provided \texttt{train} data to new \texttt{train/val}, and evaluate on unseen original \texttt{val} data, further called \texttt{test}.

\textbf{Bosch Small Traffic Lights Dataset (\texttt{BSTLD})}~\cite{behrendt2017deep}: we split the original \texttt{train} to new \texttt{train/val} with a ratio 85:15. The number of \texttt{test} images (8,334) surpasses that of \texttt{train/val} (5,093). There are 13 classes in the \texttt{train}, but only four in the \texttt{test} data (see Table~\ref{tab:classes}). We only use four classes for unified evaluation. 

\textbf{\texttt{LISA}}~\cite{jensen2016vision}: 
we split the original \texttt{train} data to new \texttt{train/val} with a ratio of 80:20 and report performance on the original \texttt{val} data.

\textbf{DriveU Traffic Light Dataset (\texttt{DTLD})}~\cite{fregin2018driveu}: 
we split train to train and val with the ratio 80:20. Similar to previous works~\cite{muller2018detecting}, we omit traffic lights belonging to classes \textit{pedestrian}, \textit{cyclist} and \textit{tram}. Each remaining traffic light contains a $1\times 1$ mapping of state and pictogram. We consider all four states related to vehicles: (\textit{red}, \textit{yellow}, \textit{red-yellow}, \textit{green}) and all five pictograms related to vehicles (\textit{circle}, \textit{arrow\_straight}, \textit{arrow\_left}, \textit{arrow\_straight\_left}, \textit{arrow\_right}). This leads to 19 possible classes, as the combination of \textit{red-yellow} state and \textit{arrow\_straight\_left} pictogram is not present in the \texttt{DTLD} dataset. Adding the class \textit{off} leads to the final 20 classes (see Table~\ref{tab:classes}).
Traffic lights labeled with \textit{unknown} states or pictograms are discarded (cf.~\cite{mentasti2023traffic}). In contrast to other works~\cite{muller2018detecting, possatti2019traffic, de2021deep}, we retain small and different-facing traffic lights in the training and test sets.

\textbf{\texttt{HDTLR}}~\cite{weber2018hdtlr}: a private dataset collected in the urban area of Karlsruhe in 2018. We use the predefined \texttt{train-val-test} split. The \texttt{train} subset was originally augmented from 3K to 43K images. After removing classes denoted as \textit{unknown}, the original hierarchy of labels results in 16 classes (see Table~\ref{tab:classes}). Turned off traffic lights as well as \textit{straight-left} and \textit{straight-right} arrows are not labeled in the dataset.

\begin{table}[h]
\begin{center}
\resizebox{1.0\linewidth}{!}{
    \begin{tabular}{|r|c| c|c|c|c| }
    \hline
    \textbf{Dataset} & \textbf{Resolution} & \textbf{Classes} & \textbf{Train} & \textbf{Val} & \textbf{Test} \\ \hline
     
    \texttt{BSTLD}~\cite{behrendt2017deep} & 1280$\times$720 & 4 & 4,329 & 764 & 8,334 \\\hline
    \texttt{LISA}~\cite{jensen2016vision} & 1280$\times$960 & 7 & 10,220 & 2,555 & 3,348 \\ \hline
    \texttt{HDTLR}~\cite{weber2018hdtlr} & 1280$\times$960 & 16 & 3,090 & 1,098 & 1,098 \\ \hline
    \texttt{DTLD}~\cite{fregin2018driveu} & 2048$\times$1024 & 20 & 22,820 & 5,705 & 12,453 \\ \hline
    \end{tabular}
}
\end{center}
\caption{Overview of the used datasets.}
\label{tab:datasets}
\end{table}

\begin{table}[t]
\centering
\resizebox{1.0\linewidth}{!}{
\begin{tabular}{|r|c|c|c|c|c|c|c|c|c|}
\hline
\textbf{Model}  &  \textbf{mAP} & \textbf{Precision} &\textbf{Recall} & 
 \makecell{\textbf{mAP}\\ \textbf{$_{3states}$}} & \makecell{\textbf{AP} \\ \textbf{$_{green}$} \\ \textbf{$_{circle}$}} & \makecell{\textbf{AP} \\ \textbf{$_{red}$} \\ \textbf{$_{circle}$}} &\makecell{\textbf{AP} \\ \textbf{$_{yellow}$} \\ \textbf{$_{circle}$}} &
 \makecell{\textbf{Speed} \\\textbf[ms]} \\ \hline


\rowcolor{LightBlue}
\multicolumn{9}{|c|}{\textbf{\texttt{BSTLD}}~\cite{behrendt2017deep}, 1280$\times$720}\\ 

YOLOv7 & \textbf{0.62} & 0.63 & \textbf{0.58} &   0.83 & 0.90 & 0.84 &0.76 & 15.7\\ 
YOLOv7x & 0.61  & 0.61 & 0.57 &  0.83 & 0.80  & 0.75  & 0.88  &  25.2 \\ 
YOLOv8m & 0.61 &0.63   & 0.52 &   0.81 & 0.88 &  0.79 & 0.77  & 18.0\\
YOLOv8x & 0.58 & 0.61  &  0.50 &   0.77 & 0.88 & 0.78 & 0.64  & 50.1\\
RT-DETR-L & 0.62 &\textbf{0.68} & 0.57 &  0.82 & 0.90 & 0.78 & 0.79 & 51.1\\
RT-DETR-X & 0.52 &0.62 &  0.50 &  0.70& 0.89 & 0.78 & 0.42 & 63.0  \\ \hline

\rowcolor{LightBlue}
\multicolumn{9}{|c|}{\textbf{\texttt{LISA}}~\cite{jensen2016vision}, 1280$\times$960}\\

YOLOv7 & \textbf{0.65} & 0.64 & \textbf{0.68} &  0.69 & 0.65 & 0.54 & 0.88  & 20.8\\
YOLOv7x & 0.64 & 0.65 & 0.63 &  0.66  &0.61 & 0.52 & 0.86  & 33.9 \\
YOLOv8m &   0.62  &0.64   & 0.58   &  0.66 & 0.57 & 0.60 & 0.81 &   18.3 \\
YOLOv8x    &  0.59  &0.67  & 0.52   &  0.71 & 0.64  & 0.64  &  0.86  & 72.7\\
RT-DETR-L & \textbf{0.65} & 0.74 & 0.57 &  0.62 & 0.53 & 0.61 & 0.73 & 52.8\\
RT-DETR-X & \textbf{0.65} &\textbf{0.80} & 0.57 &  0.52& 0.45 & 0.49 & 0.61 & 132.8\\\hline

\rowcolor{LightBlue}
\multicolumn{9}{|c|}{\textbf{\texttt{HDTLR}}~\cite{weber2018hdtlr}, 1280$\times$960 }\\ 
YOLOv7 &  0.85  & 0.81 & \textbf{0.99}  &  0.98  & 0.99  &0.97  &  0.99  & 18.1 \\
YOLOv7x & 0.86  &0.82 & \textbf{0.99}  &  0.98 & 0.99  & 0.98  & 0.99  & 30.6   \\
YOLOv8m &  \textbf{0.99}  &0.96 & 0.98  &   0.98 & 0.99   &  0.95  & 0.99   & 26.4\\
YOLOv8x    & \textbf{0.99} &0.96 & 0.98 &  0.98 & 0.99  & 0.96  & 0.99   & 71.1  \\
RT-DETR-L & 0.96 & \textbf{0.98} & 0.98 & 0.96 &0.96  & 0.92 & 0.99& 50.8\\
RT-DETR-X & 0.97 &0.95 & 0.97 &  0.94 & 0.93 & 0.93& 0.97 & 90.3\\\hline

\rowcolor{LightBlue}
\multicolumn{9}{|c|}{\textbf{\texttt{DTLD}}~\cite{fregin2018driveu}, 2048$\times$1024}\\ 
YOLOv7  & 0.59  & 0.77 & \textbf{0.54}  & 0.74 & 0.87  & 0.81 & 0.81 &24.1\\ 
YOLOv7x  & 0.60 & \textbf{0.81}  &0.53  & 0.73 & 0.87 & 0.81 & 0.81 &41.8\\  
YOLOv8m  &0.58  & 0.73 & 0.50  & 0.72   &0.85   &0.76    &0.78   & 19.8\\ 
YOLOv8x    & \textbf{0.61}&0.75 & \textbf{0.54} &    0.72 &  0.85 & 0.76   & 0.79  & 50.4\\ 
RT-DETR-X& 0.26 & 0.50& 0.27& 0.60& 0.81& 0.74& 0.63& 95.8\\ \hline 
\end{tabular}
}
\caption{Model performance. Inference speed is for a batch of size 1 on NVIDIA GeForce GTX 2080 Ti averaged over 10K predictions. YOLOv8 pads $a\times b$ images into square $a \times a$ images. Between datasets, the number of filters representing the number of classes changes for each model}
\label{tab:clean_results}
\end{table}

\begin{table*}[t]
\centering
\begin{tabular}{| r|c|c|c|c|c|c|c|c|c|c|c|c|c|c|c|c|c|c|c|c|c|}
\hline
\textbf{Dataset / Class} & \mycircle[teal] & \mycircle[red] & \mycircle[yellow] & \mycircle[orange] & off & \arrowup[teal] & \arrowup[red] & \arrowup[yellow] & \arrowup[orange] & \arrowleft[teal] & \arrowleft[red] & \arrowleft[yellow] & \arrowleft[orange] & \arrowright[teal] & \arrowright[red] & \arrowright[yellow] & \arrowright[orange] & \arrowstraightleft[teal] & \arrowstraightleft[red] & \arrowstraightleft[yellow] & \arrowstraightleft[orange] \\\hline

\texttt{BSTLD} & \cellcolor{lightgray} & \cellcolor{lightgray}& \cellcolor{lightgray}& & \cellcolor{lightgray} & & & & & & & & & & & & & & & & \\ \hline

\texttt{LISA} & \cellcolor{lightgray}& \cellcolor{lightgray}& \cellcolor{lightgray}& & & \cellcolor{lightgray}& & & & \cellcolor{lightgray}& \cellcolor{lightgray}& \cellcolor{lightgray}& & & & & & & & & \\ \hline

\texttt{HDTLR} & \cellcolor{lightgray}& \cellcolor{lightgray}& \cellcolor{lightgray}& \cellcolor{lightgray} & & \cellcolor{lightgray}& \cellcolor{lightgray}& \cellcolor{lightgray}& \cellcolor{lightgray}& \cellcolor{lightgray}& \cellcolor{lightgray}& \cellcolor{lightgray}& \cellcolor{lightgray}& \cellcolor{lightgray} & \cellcolor{lightgray}& \cellcolor{lightgray}& \cellcolor{lightgray}& & & & \\ \hline

\texttt{DTLD} & \cellcolor{lightgray}& \cellcolor{lightgray}& \cellcolor{lightgray}& \cellcolor{lightgray}& \cellcolor{lightgray}& \cellcolor{lightgray}& \cellcolor{lightgray}& \cellcolor{lightgray}& \cellcolor{lightgray}& \cellcolor{lightgray}& \cellcolor{lightgray}& \cellcolor{lightgray}& \cellcolor{lightgray}& \cellcolor{lightgray}& \cellcolor{lightgray}& \cellcolor{lightgray}& \cellcolor{lightgray}& \cellcolor{lightgray}& \cellcolor{lightgray}& \cellcolor{lightgray}& \\ \hline
\end{tabular}
\caption{Traffic light classes, available in different datasets. Orange represents yellow-red.}
\label{tab:classes}
\end{table*}

\section{MODEL EVALUATION ON VARIETY OF DATASETS}

Due to the unavailability of modern open-source implementations of traffic light detectors and especially fragmented evidence on the performance on various datasets, we first trained several generic object detectors on public and private datasets. In the following, we describe our experiments and discuss the performance on the selected datasets.

\subsection{Training Setup for Detection Task}

We use open-source PyTorch implementations of YOLOv7~\cite{wang2023yolov7}\footnote{https://github.com/WongKinYiu/yolov7}, YOLOv8~\cite{jocher2023yolov8}\footnote{https://github.com/ultralytics/ultralytics} and RT-DETR~\cite{lv2023detr}. For \texttt{BSTLD}, \texttt{LISA} and \texttt{HDTLR} datasets, we used standard model architectures, trained either on NVIDIA GeForce GTX 1080 Ti or NVIDIA GeForce RTX 2080 Ti. For \texttt{DTLD}, we modify YOLOv8 by adding the feature maps of the second stage to the Feature Pyramid Network. This increased feature map resolution improves the detection of small traffic lights. Further, we use two classification heads: one dedicated to determining states and the other for identifying pictograms. This is done to decouple the more challenging task of predicting pictograms with the easier task of color recognition. All \texttt{DTLD} models were trained for 100 epochs on 2x NVIDIA RTX 4090 or 1x NVIDIA H100 to accommodate the dataset's higher resolution and larger volume of data. For evaluation, we use precision, recall, and mean Average Precision at 50\% (mAP50) as standardized by the COCO dataset~\cite{lin2014microsoft}. Similarly standardized, all models use a confidence threshold of 0.001 and a maximum of 300 detections per image to calculate these metrics. To decouple the performance of the traffic light states from that of the pictograms, we additionally report mAP$_{3states}$, measured as a mean of AP on classes \textit{red, yellow, green}. For this, we do not retrain the models but remove the pictograms from test labels and model outputs.

\subsection{Evaluation of Detection Performance}
The evaluation (see Table~\ref{tab:clean_results}) indicates that YOLOv7 and YOLOv8 exhibit similar performance metrics. RT-DETR shows good performances on \texttt{BSTLD, LISA, HDTLR} but performs significantly worse on \texttt{DTLD}. Furthermore, the model capacity varies across the models:  YOLOv7x (70M parameters), YOLOv8x (68M parameters), and RT-DETR-X (67M parameters, 232.3 GFLOPs) are larger than YOLOv7 (36M parameters), YOLOv8m (25M parameters), and RT-DETR-L (32M parameters, 108.0 GFLOPs) resulting in longer inference times. Weber~\cite{Weber2023_1000158307} estimates that a maximum latency of 100 milliseconds is required to achieve safe braking under adverse conditions.

Direct comparison to previous works on \texttt{BSTLD} and \texttt{LISA} datasets is difficult since we deliberately decided to evaluate on data unseen during training. For both datasets, most models reach up to 0.95 mAP on validation data, thus yielding results similar to existing work. However, when evaluated on unseen data, they achieve $\leq$0.64 mAP.
Chuang~\cite{chuang2023traffic} reach with a YOLOv7 model on \texttt{BSTLD} an mAP value of 0.66, but perform no real-time performance evaluation. Their E-ELAN architecture is approximately four to five times slower than our YOLOv7x architecture \cite{wang2023yolov7}, suggesting a runtime of \textgreater 100ms, making the model unsuitable for real-time deployment. Moreover, they do not differentiate between test and validation sets as is done in our work.

For \texttt{HDTLR}, Weber et al.~\cite{weber2018hdtlr} used precision, recall, and F1 at two hierarchy levels (states and pictograms).  The best-performing model in~\cite{weber2018hdtlr} reaches a precision of 0.92 and recall of 0.96 for the resolution $1280\times960$, and correspondingly 0.84 and 0.87 for the resolution $640\times480$. By these parameters, YOLOv8 and RT-DETR models outperform those by Weber et al. Increased scores on \texttt{HDTLR} compared to other datasets can be attributed to its smaller and easier \texttt{test} dataset in which traffic lights are positioned close to the vehicle, resulting in large bounding boxes.

So far, no conventional evaluation setup for \texttt{DTLD} exists. Trinci et al.~\cite{trinci2023cross} use AP but only classify the existence of red traffic lights and do not detect bounding boxes. Müller and  Dietmayer~\cite{muller2018detecting} measure the recall for the traffic light vs. background detection and LAMR for the four predicted states. No precision values are provided, and they ignore false positives regarding pedestrian and tram traffic lights. Possatti et al.~\cite{possatti2019traffic} evaluate only AP of red and green states, additionally filtering the test dataset from 22,990 green and 16,394 red traffic lights down to 14,403 green and 9,147 red traffic lights. Bach et al.~\cite{bach2018deep} exclude labels of small, far away traffic lights, remove lesser used pictograms, and consequently reduce the test-split from originally 12,453 images to 6,653 images. In summary, to the best of our knowledge, none of the existing works using \texttt{DTLD} evaluate the detection performance of all traffic light states and pictograms.

\begin{figure}[t]
\centering
\begin{subfigure}[t]{\linewidth}
    \includegraphics[width=\textwidth]{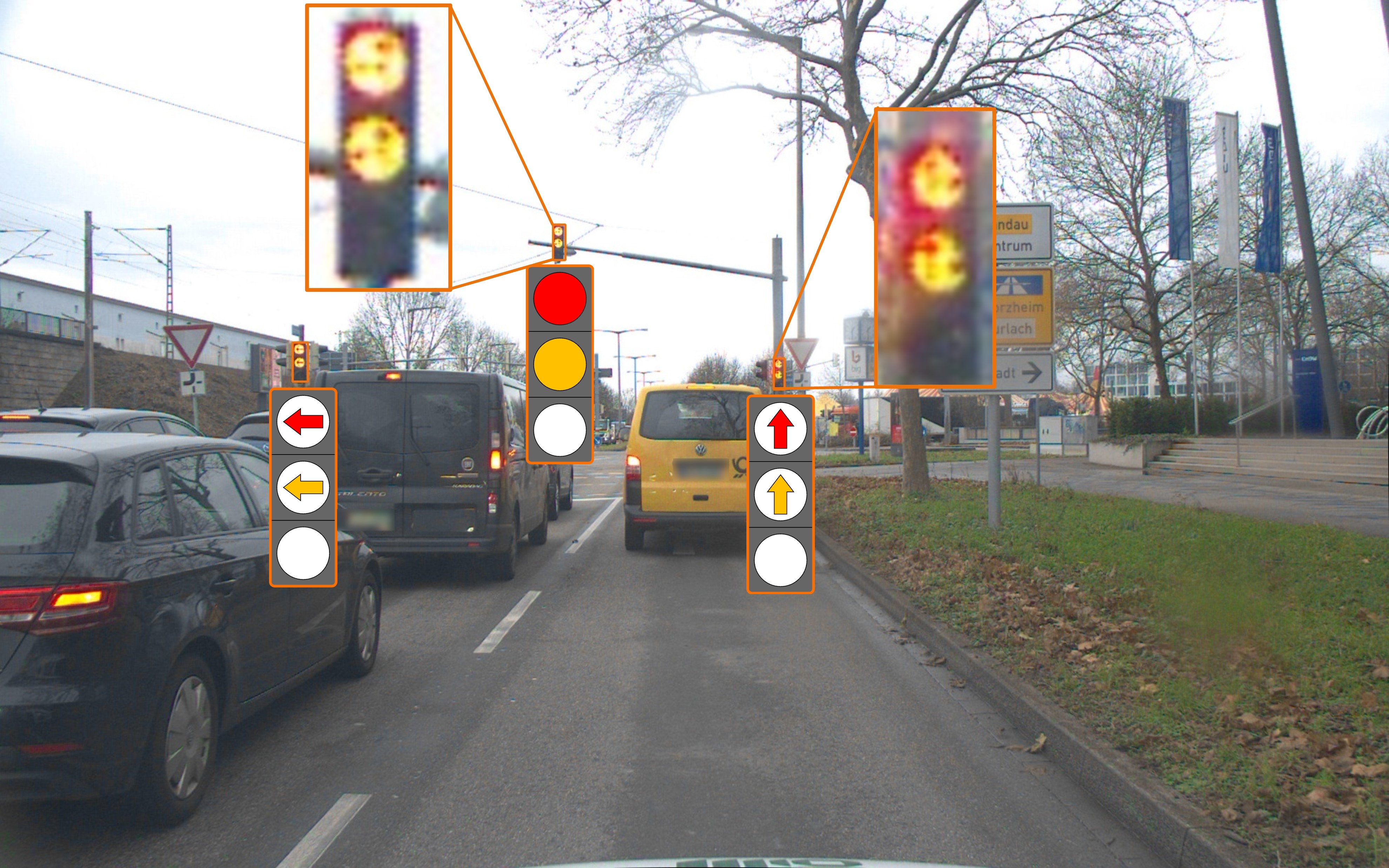}
\end{subfigure}
    \caption{Since the \texttt{DTLD} dataset does not include images of the class \textit{arrow\_straight\_right red-yellow} or the pictogram \textit{arrow\_straight\_left}, it erroneously classifies the pictograms as \textit{circle} and \textit{arrow\_straight}.}
    \label{fig:dtld-detections}
\end{figure}

\section{RELEVANCE ESTIMATION}
We propose a scalable and generalizable approach to assign relevances to traffic lights, designed to operate independently of prior maps or ground truths. Relevance estimation is crucial in complex traffic scenarios that involve multiple lanes, each with its own set of traffic lights. Therefore, we reshape the task by predicting the relevance of each lane. Then, all detected traffic lights are mapped to the lanes, whereas one lane can contain multiple traffic lights, which might not necessarily be mounted directly above.

Our process consists of three steps: 
\begin{enumerate}
\item Detection of directional arrow markings to accurately identify lanes 
\item Classification of each arrow's relevance to the ego vehicle
\item Mapping of directional arrows to corresponding traffic light pictograms, assigning relevances to ensure accurate relevance estimation
\end{enumerate}
We hypothesize that detecting and assigning relevance to directional arrows is more feasible than directly estimating the relevance of traffic lights, primarily due to their larger size and greater visibility.
Furthermore, the positioning of relevant arrow markings typically close to the center of the image simplifies the training process.
For this approach, we annotate road marking with bounding boxes and classes, \textit{straight, left, right, straight\_left, straight\_right}, across approximately 11,000 training images and approximately 3,000 images for testing from the \texttt{DTLD} dataset. Additionally, we label the relevance of arrows for the ego vehicle. For reproducibility and providing a foundation for further research, we also open-source these newly created labels.
First, we train a new object detection model for detecting directional arrow markings. As these arrows are relatively large, we employ a smaller YOLV8m model with a starting learning rate of 0.01, which is cosine-annealed over 50 epochs.
\begin{table}
\begin{center}
\resizebox{1.0\linewidth}{!}{
    \begin{tabular}{|r|c| c|c|c|}
    \hline
    \rowcolor{YellowGreen}
 \textbf{Directional Arrows } & \textbf{No. Instances} & \textbf{Precision} & \textbf{Recall} & \textbf{mAP}\\ \hline

     \rowcolor{lightgray}\textbf{all} & \textbf{9276} & \textbf{0.92}& \textbf{0.9} & \textbf{0.94} \\ \hline
    Straight & 4507 & 0.92& 0.92 &0.95 \\ \hline
    Left & 2237 & 0.93& 0.96 &0.97 \\ \hline
    Right & 814 & 0.90& 0.90 &0.94\\ \hline
    Straight-Right & 288 & 0.93& 0.83 &0.90\\ \hline
    Straight-Left & 1430 & 0.96& 0.88 &0.94\\ \hline
    \end{tabular}
}

\end{center}
\caption{Evaluation of the directional arrow marking detection on the test dataset}
\label{tab:arrowmarking}
\end{table}
The model's performance, as evaluated in Table~\ref{tab:arrowmarking}, remains robust even under conditions of damaged roads or faded markings. Manual inspection of incorrect predictions reveals that false positives are predominantly caused by misinterpretations of other road markings.

In the second step, to predict the relevance of each directional arrow, we train a gradient boosting method~\cite{friedman2002stochastic} using 300 boosting stages and a maximum depth of three. It takes the predictions of the directional arrow-marking detection model as input:
\begin{itemize}
    \item Predicted center coordinates of bounding box
    \item Width and height of bounding box
    \item Classification for the type of the directional arrow
    \item Deviation from the center of the image
\end{itemize}
and returns for each arrow a binary classification into \mbox{\textit{relevant}} / \mbox{\textit{not-relevant}}. Table~\ref{tab:relevance} presents the evaluation on a hold-out test dataset.

\begin{figure}[t]
\centering
    \includegraphics[width=\linewidth]{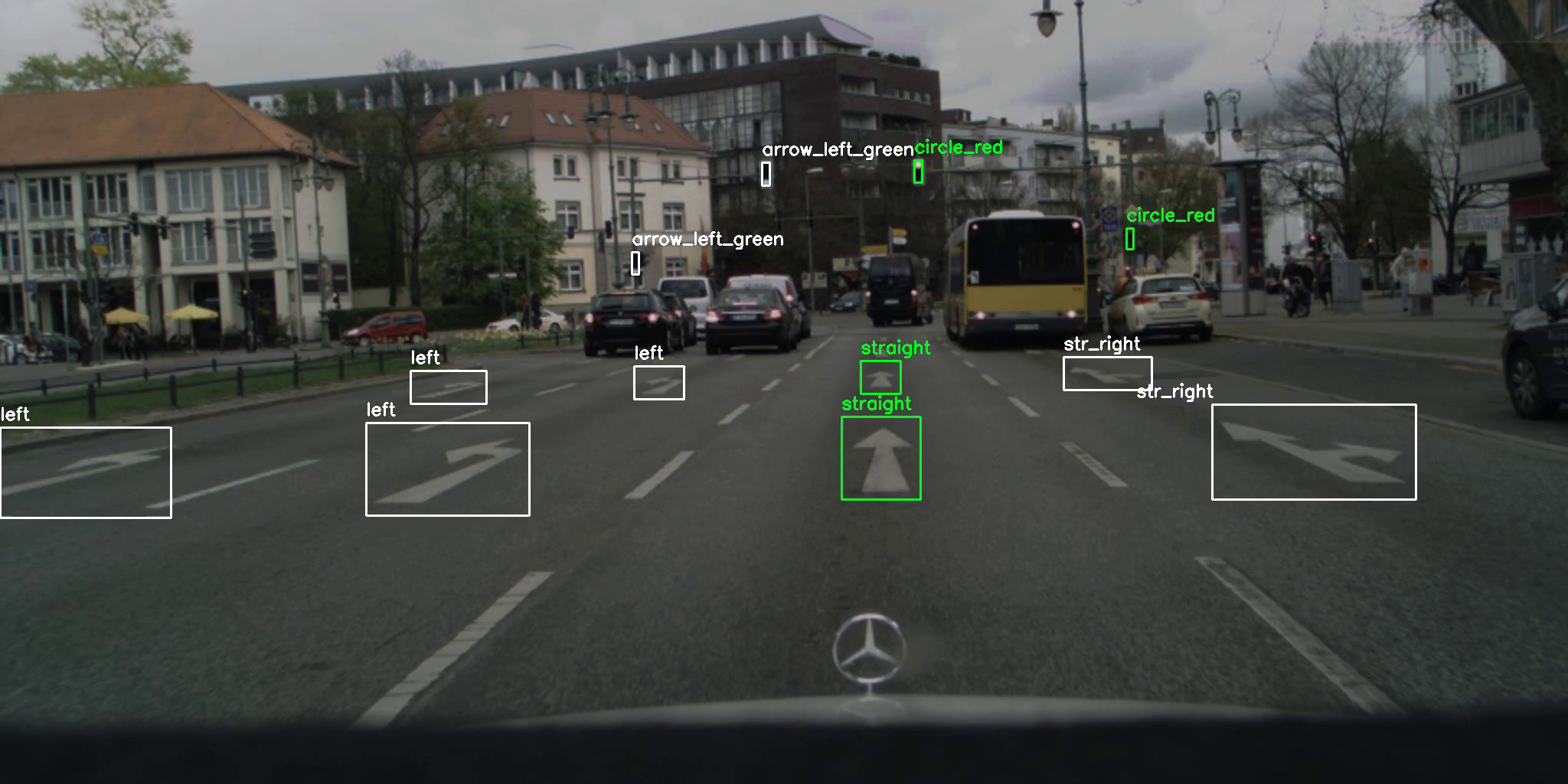}
    \caption{Detection of arrow markings and traffic lights on \texttt{DTLD} dataset.
    Relevant directional arrows are marked in green. All traffic lights matched to these relevant arrows are also marked in green. Non-relevant arrows and traffic lights are kept in white. The rightmost traffic light is relevant, as in this situation, the \textit{circle} pictogram encompasses the straight arrow}
    \label{fig:relevance}
\end{figure}

In the third step, we establish a mapping between detected traffic light pictograms and detected directional arrows. The detected pictograms are compared to the detected directional arrows, and a best-fit matching algorithm is applied. E.g., a \textit{right} directional arrow is generally mapped to a \textit{right} pictogram, but in its absence, it is mapped to the \textit{circle} pictogram. The relevance score of the directional arrow is then assigned to all mapped traffic lights. Figure~\ref{fig:relevance} illustrates the accumulation of these steps, highlighting relevant directional arrows and traffic lights in green. A notable challenge of this approach arises in scenarios in which no directional arrows are visible. This usually happens if the vehicle is close to the stopping line of the traffic lights or the view of the markings is blocked. To combat this, a sliding window historical record of earlier predicted relevant directional arrows is kept and applied in these cases. In the case of single lanes in which all traffic lights contain the same pictogram, all traffic lights are relevant. Additionally, in rare cases in which all traffic lights show the same state, relevance estimation is unnecessary, and all are assigned as relevant. 
To conclude, our three-step model integrates directional arrow detection with traffic light relevance, offering a solution suited to environments lacking prior maps. 

\begin{table}[h]
\begin{center}
\resizebox{1.0\linewidth}{!}{
    \begin{tabular}{|r|c| c|c|c|}
    \hline
    \rowcolor{YellowGreen}
    \textbf{Method } & \textbf{Precision} & \textbf{Recall} & \textbf{Accuracy}\\ \hline
     \rowcolor{lightgray}\textbf{Gradient Boosting} & \textbf{0.96} & \textbf{0.96}& \textbf{0.96}\\ \hline
    \end{tabular}
}

\end{center}
\caption{Evaluation of relevance estimation on test dataset}
\label{tab:relevance}
\end{table}

\section{REAL-WORLD EVALUATION}

\begin{figure}[b]
\centering
    \includegraphics[width=\linewidth]{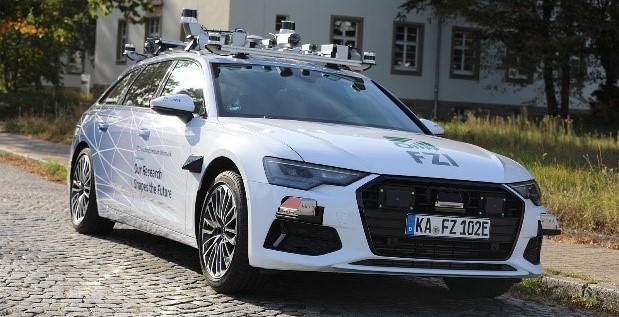}
    \caption{Research vehicle \textit{CoCar NextGen} used for test drives.}
    \label{fig:ccng}
\end{figure}

In the following, we describe how models trained on the datasets perform during test drives in unseen urban areas.

To evaluate the generalization and real-time abilities of the trained detectors, we deploy them in our research vehicle \textit{CoCar NextGen}~\cite{heinrich2024cocarnextgen}, which is based on an Audi Q5 and equipped with LiDAR, GNSS + IMU, and RGB camera systems, see Figure~\ref{fig:ccng}. 
We use the frontal medium central camera to get rectified images at 40 FPS with a resolution of $1920\times1200$. We deploy the YOLOv8x model, trained on \texttt{DTLD} on the onboard server, using an Nvidia RTX A6000. The Robot Operating System (ROS)~\cite{Quigley09} is used as the communication protocol between the camera and model. For real-time human monitoring, the model's traffic light predictions are visually showcased on the vehicle's displays. We further evaluate the generalizability of the model across different cameras, lenses, and driving environments using ten separate test rides in the urban area of Karlsruhe, Germany. The full end-to-end latency in the vehicle, including model inference, image pre- and post-processing, in-vehicle networking via ROS, and visualization on the vehicle's monitors, requires an average of 52ms per image with an average 60\% GPU-utilization. Predictions of traffic light states are almost always accurate, with misclassifications of pictograms occurring primarily at larger distances. Figure~\ref{fig:concept} demonstrates the detection capabilities in various scenarios. Notably, we observed a significant number of false positive detections, particularly in scenarios in which no traffic lights are visible. Other challenges persist with detecting classes and pictograms that are not represented in any dataset. Figure~\ref{fig:dtld-detections} illustrates an urban traffic scene containing traffic lights with \textit{straight-left} and \textit{straight-right} pictograms, which the model fails to detect due to these limitations.
We manually annotate 9,600 images from these test drives to quantify and fine-tune the model. This dataset, which also includes \texttt{straight\_right} and \texttt{straight\_left} pictograms, is split into 9,000 images for training and validation and 600 images from a separate test drive conducted in a different season for testing. The model, without any modifications, achieves an mAP of 0.28 on this new dataset. However, fine-tuning the model significantly increases the mAP to 0.959.

Another challenge is the relevance estimation. While our directional arrow detection model demonstrates robust performance on real-world data, the relevance classification model struggles with adapting to the new data format, yielding insufficient predictions. Given that its input is of a tabular nature, fine-tuning using custom data is necessary. 

\section{CONCLUSION}
We propose a comprehensive approach to traffic light detection and relevance estimation for real-time utilization in autonomous vehicles. By comparing three modern model architectures on four separate datasets, we create a foundation for successful traffic light detection. By using standardized metrics and not removing images from test datasets, our results are easily comparable to future work. A novel relevance estimation method, leveraging directional arrow markings instead of relying on costly pre-annotated maps, achieves high accuracies on the test dataset of \texttt{DTLD}. The real-world test drives evaluate the robustness of the methodologies to new data and ascertain real-time performance. We contribute to the research community by open-sourcing our models, code, and the newly created dataset for relevance estimation on \texttt{DTLD}. An unsolved challenge is the lack of comprehensive datasets containing all relevant classes.  

\vfill

\section*{Acknowledgment}

This work was supported by funding from the Topic Engineering Secure Systems of the Helmholtz
Association (HGF) and by KASTEL Security Research Labs (46.23.03).
\newpage

{\small
\bibliographystyle{IEEEtran}
\bibliography{references}

\begin{thebibliography}{10}
\providecommand{\url}[1]{#1}
\csname url@samestyle\endcsname
\providecommand{\newblock}{\relax}
\providecommand{\bibinfo}[2]{#2}
\providecommand{\BIBentrySTDinterwordspacing}{\spaceskip=0pt\relax}
\providecommand{\BIBentryALTinterwordstretchfactor}{4}
\providecommand{\BIBentryALTinterwordspacing}{\spaceskip=\fontdimen2\font plus
\BIBentryALTinterwordstretchfactor\fontdimen3\font minus \fontdimen4\font\relax}
\providecommand{\BIBforeignlanguage}[2]{{%
\expandafter\ifx\csname l@#1\endcsname\relax
\typeout{** WARNING: IEEEtran.bst: No hyphenation pattern has been}%
\typeout{** loaded for the language `#1'. Using the pattern for}%
\typeout{** the default language instead.}%
\else
\language=\csname l@#1\endcsname
\fi
#2}}
\providecommand{\BIBdecl}{\relax}
\BIBdecl

\bibitem{ochs2024one}
S.~Ochs, J.~Doll, D.~Grimm, T.~Fleck, M.~Heinrich, S.~Orf, A.~Schotschneider, H.~Gremmelmaier, R.~Polley, S.~Pavlitska \emph{et~al.}, ``{One Stack to Rule them All: To Drive Automated Vehicles, and Reach for the 4th level},'' \emph{arXiv preprint arXiv:2404.02645}, 2024.

\bibitem{zipfl2020traffic}
M.~Zipfl, T.~Fleck, M.~R. Zofka, and J.~M. Z{\"o}llner, ``From traffic sensor data to semantic traffic descriptions: The test area autonomous driving baden-w{\"u}rttemberg dataset (taf-bw dataset),'' in \emph{International Conference on Intelligent Transportation Systems (ITSC)}.\hskip 1em plus 0.5em minus 0.4em\relax IEEE, 2020.

\bibitem{pavlitska2023traffic}
S.~Pavlitska, N.~Lambing, A.~Kimar~Bangaru, and J.~M. Z{\"o}llner, ``{Traffic Light Recognition using Convolutional Neural Networks: A Survey},'' in \emph{International Conference on Intelligent Transportation Systems (ITSC)}, 2023.

\bibitem{chung2002vision}
Y.-C. Chung, J.-M. Wang, and S.-W. Chen, ``A vision-based traffic light detection system at intersections,'' \emph{Journal of Taiwan Normal University: Mathematics, Science and Technology}, 2002.

\bibitem{lindner2004robust}
F.~Lindner, U.~Kressel, and S.~Kaelberer, ``Robust recognition of traffic signals,'' in \emph{Intelligent Vehicles Symposium (IV)}.\hskip 1em plus 0.5em minus 0.4em\relax IEEE, 2004.

\bibitem{nienhuser2010visual}
D.~Nienh{\"u}ser, M.~Drescher, and J.~M. Z{\"o}llner, ``Visual state estimation of traffic lights using hidden markov models,'' in \emph{International Conference on Intelligent Transportation Systems (ITSC)}.\hskip 1em plus 0.5em minus 0.4em\relax IEEE, 2010.

\bibitem{john2014traffic}
V.~John, K.~Yoneda, B.~Qi, Z.~Liu, and S.~Mita, ``Traffic light recognition in varying illumination using deep learning and saliency map,'' in \emph{International Conference on Intelligent Transportation Systems (ITSC)}.\hskip 1em plus 0.5em minus 0.4em\relax {IEEE}, 2014.

\bibitem{john2015saliency}
V.~John, K.~Yoneda, Z.~Liu, and S.~Mita, ``Saliency map generation by the convolutional neural network for real-time traffic light detection using template matching,'' \emph{{IEEE} Trans. Computational Imaging}, 2015.

\bibitem{possatti2019traffic}
{L. C. Possatti, R. Guidolini, V. Cardoso, et al.}, ``Traffic light recognition using deep learning and prior maps for autonomous cars,'' in \emph{International Joint Conference on Neural Networks (IJCNN)}.\hskip 1em plus 0.5em minus 0.4em\relax IEEE, 2019.

\bibitem{sanitz2023small}
T.~Sanitz, C.~Wilms, and S.~Frintrop, ``Small, but important: Traffic light proposals for detecting small traffic lights and beyond,'' in \emph{International Conference on Computer Vision Systems}.\hskip 1em plus 0.5em minus 0.4em\relax Springer, 2023.

\bibitem{redmon2016you}
J.~Redmon, S.~Divvala, R.~Girshick, and A.~Farhadi, ``You only look once: Unified, real-time object detection,'' in \emph{Conference on Computer Vision and Pattern Recognition (CVPR)}, 2016.

\bibitem{redmon2018yolo9000}
J.~Redmon and A.~Farhadi, ``{YOLO9000:} better, faster, stronger,'' in \emph{Conference on Computer Vision and Pattern Recognition (CVPR)}.\hskip 1em plus 0.5em minus 0.4em\relax {IEEE} Computer Society, 2017.

\bibitem{redmon2018yolov3}
{J. Redmon and A. Farhadi}, ``Yolov3: An incremental improvement,'' in \emph{arXiv preprint arXiv:1804.02767}, 2018.

\bibitem{bochkovskiy2020yolov4}
A.~Bochkovskiy, C.~Wang, and H.~M. Liao, ``Yolov4: Optimal speed and accuracy of object detection,'' \emph{CoRR}, vol. abs/2004.10934, 2020.

\bibitem{liu2016ssd}
W.~Liu, D.~Anguelov, D.~Erhan, C.~Szegedy, S.~Reed, C.-Y. Fu, and A.~C. Berg, ``Ssd: Single shot multibox detector,'' in \emph{European Conference on Computer Vision (ECCV)}.\hskip 1em plus 0.5em minus 0.4em\relax Springer, 2016.

\bibitem{ren2015faster}
S.~Ren, K.~He, R.~B. Girshick, and J.~Sun, ``Faster {R-CNN:} towards real-time object detection with region proposal networks,'' in \emph{Advances in Neural Information Processing Systems (NIPS)}, 2015.

\bibitem{aneesh2019real}
A.~Aneesh, L.~Shine, R.~Pradeep, and V.~Sajith, ``Real-time traffic light detection and recognition based on deep retinanet for self driving cars,'' in \emph{International Conference on Intelligent Computing, Instrumentation and Control Technologies (ICICICT)}.\hskip 1em plus 0.5em minus 0.4em\relax IEEE, 2019.

\bibitem{bach2018deep}
M.~Bach, D.~Stumper, and K.~Dietmayer, ``Deep convolutional traffic light recognition for automated driving,'' in \emph{International Conference on Intelligent Transportation Systems (ITSC)}.\hskip 1em plus 0.5em minus 0.4em\relax IEEE, 2018.

\bibitem{gokul2020comparative}
R.~Gokul, A.~Nirmal, K.~Bharath, M.~Pranesh, and R.~Karthika, ``A comparative study between state-of-the-art object detectors for traffic light detection,'' in \emph{International Conference on Emerging Trends in Information Technology and Engineering (ic-ETITE)}.\hskip 1em plus 0.5em minus 0.4em\relax IEEE, 2020.

\bibitem{jensen2016vision}
M.~B. Jensen, M.~P. Philipsen, A.~M{\o}gelmose, T.~B. Moeslund, and M.~M. Trivedi, ``Vision for looking at traffic lights: Issues, survey, and perspectives,'' \emph{IEEE transactions on intelligent transportation systems}, 2016.

\bibitem{liu2023traffic}
P.~Liu and T.~Li, ``Traffic light detection based on depth improved yolov5,'' in \emph{International Conference on Neural Networks, Information and Communication Engineering (NNICE)}.\hskip 1em plus 0.5em minus 0.4em\relax IEEE, 2023.

\bibitem{muller2018detecting}
J.~M{\"u}ller and K.~Dietmayer, ``Detecting traffic lights by single shot detection,'' in \emph{International Conference on Intelligent Transportation Systems (ITSC)}.\hskip 1em plus 0.5em minus 0.4em\relax IEEE, 2018.

\bibitem{pon2018hierarchical}
A.~Pon, O.~Adrienko, A.~Harakeh, and S.~L. Waslander, ``A hierarchical deep architecture and mini-batch selection method for joint traffic sign and light detection,'' in \emph{Conference on Computer and Robot Vision (CRV)}.\hskip 1em plus 0.5em minus 0.4em\relax IEEE, 2018.

\bibitem{yan2021end}
S.~Yan, X.~Liu, W.~Qian, and Q.~Chen, ``An end-to-end traffic light detection algorithm based on deep learning,'' in \emph{International conference on security, pattern analysis, and cybernetics (SPAC)}.\hskip 1em plus 0.5em minus 0.4em\relax IEEE, 2021.

\bibitem{han2019real}
C.~Han, G.~Gao, and Y.~Zhang, ``Real-time small traffic sign detection with revised faster-rcnn,'' \emph{Multimedia Tools and Applications}, 2019.

\bibitem{wang2022traffic}
Q.~Wang, Q.~Zhang, X.~Liang, Y.~Wang, C.~Zhou, and V.~I. Mikulovich, ``Traffic lights detection and recognition method based on the improved yolov4 algorithm,'' \emph{Sensors}, 2022.

\bibitem{naimi2021fast}
H.~Naimi, T.~Akilan, and M.~A. Khalid, ``Fast traffic sign and light detection using deep learning for automotive applications,'' in \emph{IEEE Western New York Image and Signal Processing Workshop (WNYISPW)}.\hskip 1em plus 0.5em minus 0.4em\relax IEEE, 2021.

\bibitem{weber2016deeptlr}
M.~Weber, P.~Wolf, and J.~M. Z{\"o}llner, ``{DeepTLR: A single deep convolutional network for detection and classification of traffic lights},'' in \emph{Intelligent Vehicles Symposium (IV)}.\hskip 1em plus 0.5em minus 0.4em\relax IEEE, 2016.

\bibitem{weber2018hdtlr}
M.~Weber, M.~Huber, and J.~M. Z{\"o}llner, ``{HDTLR: A CNN based hierarchical detector for traffic lights},'' in \emph{International Conference on Intelligent Transportation Systems (ITSC)}.\hskip 1em plus 0.5em minus 0.4em\relax IEEE, 2018.

\bibitem{lu2018traffic}
Y.~Lu, J.~Lu, S.~Zhang, and P.~Hall, ``Traffic signal detection and classification in street views using an attention model,'' \emph{Computational Visual Media}, 2018.

\bibitem{wang2018method}
X.~Wang, T.~Jiang, and Y.~Xie, ``A method of traffic light status recognition based on deep learning,'' in \emph{International Conference on Robotics, Control and Automation Engineering}, 2018.

\bibitem{kim2019traffic}
H.-K. Kim, K.-Y. Yoo, J.~H. Park, and H.-Y. Jung, ``Traffic light recognition based on binary semantic segmentation network,'' \emph{Sensors}, 2019.

\bibitem{jayasinghe2022towards}
O.~Jayasinghe, S.~Hemachandra, D.~Anhettigama, S.~Kariyawasam, T.~Wickremasinghe, C.~Ekanayake, R.~Rodrigo, and P.~Jayasekara, ``Towards real-time traffic sign and traffic light detection on embedded systems,'' in \emph{Intelligent Vehicles Symposium (IV)}.\hskip 1em plus 0.5em minus 0.4em\relax IEEE, 2022.

\bibitem{choi2024real}
J.~Choi and H.~Lee, ``Real-time traffic light recognition with lightweight state recognition and ratio-preserving zero padding,'' \emph{Electronics}, 2024.

\bibitem{kim2018deep}
J.~Kim, H.~Cho, M.~Hwangbo, J.~Choi, J.~Canny, and Y.~P. Kwon, ``Deep traffic light detection for self-driving cars from a large-scale dataset,'' in \emph{International Conference on Intelligent Transportation Systems (ITSC)}.\hskip 1em plus 0.5em minus 0.4em\relax IEEE, 2018.

\bibitem{yudin2018usage}
D.~Yudin and D.~Slavioglo, ``Usage of fully convolutional network with clustering for traffic light detection,'' in \emph{Mediterranean Conference on Embedded Computing (MECO)}.\hskip 1em plus 0.5em minus 0.4em\relax IEEE, 2018.

\bibitem{gupta2019framework}
A.~Gupta and A.~Choudhary, ``A framework for traffic light detection and recognition using deep learning and grassmann manifolds,'' in \emph{Intelligent Vehicles Symposium (IV)}.\hskip 1em plus 0.5em minus 0.4em\relax IEEE, 2019.

\bibitem{tran2020accurate}
T.~H.-P. Tran and J.~W. Jeon, ``Accurate real-time traffic light detection using yolov4,'' in \emph{IEEE International Conference on Consumer Electronics-Asia (ICCE-Asia)}.\hskip 1em plus 0.5em minus 0.4em\relax IEEE, 2020.

\bibitem{nguyen2020robust}
P.~M. Nguyen, V.~C. Nguyen, S.~N. Nguyen, L.~M.~T. Dang, H.~X. Nguyen, and V.~D. Nguyen, ``Robust traffic light detection and classification under day and night conditions,'' in \emph{International conference on control, automation and systems (ICCAS)}.\hskip 1em plus 0.5em minus 0.4em\relax IEEE, 2020.

\bibitem{zhang2023robust}
Y.~Zhang, Y.~Wu, J.~Zhao, Y.~Hu, Y.~He, and J.~Wang, ``Robust traffic light recognition pipeline based on yolov8 for autonomous driving systems,'' in \emph{2023 IEEE 29th International Conference on Parallel and Distributed Systems (ICPADS)}.\hskip 1em plus 0.5em minus 0.4em\relax IEEE, 2023, pp. 1278--1285.

\bibitem{jocher2023yolov8}
G.~Jocher, A.~Chaurasia, and J.~Qiu, ``Yolo by ultralytics (version 8.0.0) [computer software],'' \url{https://github.com/ultralytics/ultralytics}, 2023.

\bibitem{vaswani2017attention}
A.~Vaswani, N.~Shazeer, N.~Parmar, J.~Uszkoreit, L.~Jones, A.~N. Gomez, {\L}.~Kaiser, and I.~Polosukhin, ``Attention is all you need,'' \emph{Advances in Neural Information Processing Systems (NIPS)}, 2017.

\bibitem{carion2020end}
N.~Carion, F.~Massa, G.~Synnaeve, N.~Usunier, A.~Kirillov, and S.~Zagoruyko, ``End-to-end object detection with transformers,'' in \emph{European Conference on Computer Vision (ECCV)}.\hskip 1em plus 0.5em minus 0.4em\relax Springer, 2020.

\bibitem{greer2023robust}
R.~Greer, A.~Gopalkrishnan, J.~Landgren, L.~Rakla, A.~Gopalan, and M.~M. Trivedi, ``Robust traffic light detection using salience-sensitive loss: Computational framework and evaluations,'' \emph{CoRR}, vol. abs/2305.04516, 2023.

\bibitem{ou2022traffic}
Y.~Ou, Y.~Sun, X.~Yu, and L.~Yun, ``Traffic signal light recognition based on transformer,'' in \emph{International Conference on Computer Engineering and Networks}.\hskip 1em plus 0.5em minus 0.4em\relax Springer, 2022.

\bibitem{chuang2023traffic}
C.-H. Chuang, C.-C. Lee, J.-H. Lo, and K.-C. Fan, ``Traffic light detection by integrating feature fusion and attention mechanism,'' \emph{Electronics}, vol.~12, no.~17, p. 3727, 2023.

\bibitem{wang2023yolov7}
C.~Wang, A.~Bochkovskiy, and H.~M. Liao, ``Yolov7: Trainable bag-of-freebies sets new state-of-the-art for real-time object detectors,'' in \emph{Conference on Computer Vision and Pattern Recognition (CVPR)}, 2023.

\bibitem{lv2023detr}
W.~Lv, S.~Xu, Y.~Zhao, G.~Wang, J.~Wei, C.~Cui, Y.~Du, Q.~Dang, and Y.~Liu, ``Detrs beat yolos on real-time object detection,'' \emph{CoRR}, vol. abs/2304.08069, 2023.

\bibitem{apollo}
``Baidu apollo team (2017), apollo: Open source autonomous driving,'' \url{https://github.com/ApolloAuto/apollo}, accessed: 2024-03-20.

\bibitem{fairfield2011traffic}
N.~Fairfield and C.~Urmson, ``Traffic light mapping and detection,'' in \emph{International Conference on Robotics and Automation (ICRA)}.\hskip 1em plus 0.5em minus 0.4em\relax IEEE, 2011.

\bibitem{levinson2011traffic}
J.~Levinson, J.~Askeland, J.~Dolson, and S.~Thrun, ``Traffic light mapping, localization, and state detection for autonomous vehicles,'' in \emph{International Conference on Robotics and Automation (ICRA)}.\hskip 1em plus 0.5em minus 0.4em\relax IEEE, 2011.

\bibitem{langenberg2019deep}
T.~Langenberg, T.~L{\"u}ddecke, and F.~W{\"o}rg{\"o}tter, ``Deep metadata fusion for traffic light to lane assignment,'' \emph{IEEE Robotics and Automation Letters}, 2019.

\bibitem{li2017traffic}
X.~Li, H.~Ma, X.~Wang, and X.~Zhang, ``Traffic light recognition for complex scene with fusion detections,'' \emph{International Conference on Intelligent Transportation Systems (ITSC)}, 2017.

\bibitem{trinci2023cross}
T.~Trinci, T.~Bianconcini, L.~Sarti, L.~Taccari, and F.~Sambo, ``Cross-model temporal cooperation via saliency maps for efficient frame classification,'' in \emph{International Conference on Computer Vision (ICCV)}, 2023.

\bibitem{LARA}
R.~de~Charette, ``Lara french traffic lights recognition (tlr) public benchmarks,'' 2015.

\bibitem{chen2016accurate}
Z.~Chen and X.~Huang, ``Accurate and reliable detection of traffic lights using multiclass learning and multiobject tracking,'' \emph{{IEEE} Intell. Transp. Syst. Mag.}, 2016.

\bibitem{janosovits2022cityscapes}
J.~Janosovits, ``Cityscapes tl++: Semantic traffic light annotations for the cityscapes dataset,'' in \emph{International Conference on Robotics and Automation (ICRA)}.\hskip 1em plus 0.5em minus 0.4em\relax IEEE, 2022.

\bibitem{yang2022scrdet++}
X.~Yang, J.~Yan, W.~Liao, X.~Yang, J.~Tang, and T.~He, ``Scrdet++: Detecting small, cluttered and rotated objects via instance-level feature denoising and rotation loss smoothing,'' \emph{IEEE Transactions on Pattern Analysis and Machine Intelligence}, 2022.

\bibitem{jayarathne2023dualcam}
{H. Jayarathne, T. Samarakoon, H. Koralege et al.}, ``Dualcam: A novel benchmark dataset for fine-grained real-time traffic light detection,'' in \emph{International Conference on Machine Learning and Applications (ICMLA)}.\hskip 1em plus 0.5em minus 0.4em\relax IEEE, 2023.

\bibitem{behrendt2017deep}
K.~Behrendt, L.~Novak, and R.~Botros, ``A deep learning approach to traffic lights: Detection, tracking, and classification,'' in \emph{International Conference on Robotics and Automation (ICRA)}.\hskip 1em plus 0.5em minus 0.4em\relax IEEE, 2017.

\bibitem{fregin2018driveu}
A.~Fregin, J.~Mueller, U.~Kreßel, and K.~Dietmayer, ``The driveu traffic light dataset: Introduction and comparison with existing datasets,'' in \emph{International Conference on Robotics and Automation (ICRA)}.\hskip 1em plus 0.5em minus 0.4em\relax IEEE, 2018.

\bibitem{mentasti2023traffic}
S.~Mentasti, Y.~C. Simsek, and M.~Matteucci, ``Traffic lights detection and tracking for {HD} map creation,'' \emph{Frontiers Robotics {AI}}, 2023.

\bibitem{de2021deep}
J.~P.~V. de~Mello, L.~Tabelini, R.~F. Berriel, T.~M. Paixao, A.~F. De~Souza, C.~Badue, N.~Sebe, and T.~Oliveira-Santos, ``Deep traffic light detection by overlaying synthetic context on arbitrary natural images,'' \emph{Computers \& Graphics}, 2021.

\bibitem{lin2014microsoft}
{T.-Y. Lin, M. Maire, S. Belongie et al.}, ``Microsoft {COCO}: Common objects in context,'' in \emph{European Conference on Computer Vision (ECCV)}.\hskip 1em plus 0.5em minus 0.4em\relax Springer, 2014.

\bibitem{Weber2023_1000158307}
M.~Weber, ``{Integrierte Konzepte tiefer Neuronaler Netze zur monokularen Informationsgewinnung im Autonomen Fahrzeug},'' Ph.D. dissertation, {Karlsruher Institut f{\"{u}}r Technologie (KIT)}, 2023.

\bibitem{friedman2002stochastic}
J.~H. Friedman, ``Stochastic gradient boosting,'' \emph{Computational statistics \& data analysis}, 2002.

\bibitem{heinrich2024cocarnextgen}
M.~Heinrich, M.~Zipfl, M.~Uecker, S.~Ochs, M.~Gontscharow, T.~Fleck, J.~Doll, P.~Schörner, C.~Hubschneider, M.~R. Zofka, A.~Viehl, and J.~M. Zöllner, ``{CoCar NextGen: a Multi-Purpose Platform for Connected Autonomous Driving Research},'' \emph{arXiv preprint arXiv:2404.17550}, 2024.

\bibitem{Quigley09}
M.~Quigley, B.~Gerkey, K.~Conley, J.~Faust, T.~Foote, J.~Leibs, E.~Berger, R.~Wheeler, and A.~Ng, ``{ROS: an open-source Robot Operating System},'' in \emph{International Conference on Robotics and Automation (ICRA)}, 2009.

\end{thebibliography}
}

\end{document}